\documentclass[conference]{IEEEtran}
\IEEEoverridecommandlockouts
\usepackage{cite}
\usepackage{amsmath,amssymb,amsfonts}
\usepackage{algorithmic}
\usepackage{graphicx}
\usepackage{textcomp}
\usepackage{subcaption}
\usepackage{longtable}
\usepackage{lscape}
\usepackage{url}
\usepackage{xcolor}
\usepackage{float}
\usepackage{array}
\usepackage{footnote}
\usepackage{parcolumns}
\usepackage{hyperref}
\usepackage{multirow}
\usepackage{booktabs}
\usepackage{hhline}
\usepackage{colortbl}
\usepackage{adjustbox}
\usepackage{rotating}

\usepackage{enumitem,booktabs}
\usepackage[referable]{threeparttablex}
\renewlist{tablenotes}{enumerate}{1}
\makeatletter
\setlist[tablenotes]{label=\tnote{\alph*},ref=\alph*,itemsep=\z@,topsep=\z@skip,partopsep=\z@skip,parsep=\z@,itemindent=\z@,labelindent=\tabcolsep,labelsep=.2em,leftmargin=*,align=left,before={\footnotesize}}
\makeatother

\def\BibTeX{{\rm B\kern-.05em{\sc i\kern-.025em b}\kern-.08em
    T\kern-.1667em\lower.7ex\hbox{E}\kern-.125emX}}
\begin{document}

\newcolumntype{L}{>{\centering\arraybackslash}m{1.8cm}}
\newcolumntype{K}{>{\centering\arraybackslash}m{3cm}}
\newcolumntype{J}{>{\centering\arraybackslash}m{2.7cm}}

\newcolumntype{A}{>{\centering\arraybackslash}m{2cm}}
\newcolumntype{B}{>{\centering\arraybackslash}m{4cm}}
\newcolumntype{C}{>{\centering\arraybackslash}m{5cm}}
\newcolumntype{D}{>{\centering\arraybackslash}m{3cm}}

\newcolumntype{Y}{>{\arraybackslash}m{5cm}}

\title{Improved RAMEN: Towards Domain Generalization for Visual Question Answering}

\author{\IEEEauthorblockN{Bhanuka Manesha Samarasekara Vitharana Gamage}
\IEEEauthorblockA{\textit{School of Information Technology} \\
\textit{Monash University}\\
Bandar Sunway, Malaysia \\
bsam0002@student.monash.edu}
\and
\IEEEauthorblockN{Lim Chern Hong}
\IEEEauthorblockA{\textit{School of Information Technology} \\
\textit{Monash University}\\
Bandar Sunway, Malaysia\\
lim.chernhong@monash.edu}
}

\maketitle

\begin{abstract}
    Currently nearing human-level performance, Visual Question Answering (VQA) is an emerging area in artificial intelligence.
    Established as a multi-disciplinary field in machine learning, both computer vision and natural language processing communities are working together to achieve state-of-the-art (SOTA) performance.
    However, there is a gap between the SOTA results and real world applications.
    This is due to the lack of model generalisation.
    The RAMEN model \cite{Shrestha2019} aimed to achieve domain generalization by obtaining the highest score across two main types of VQA datasets.
    This study provides two major improvements to the early/late fusion module and aggregation module of the RAMEN architecture, with the objective of further strengthening domain generalization.
    Vector operations based fusion strategies are introduced for the fusion module and the transformer architecture is introduced for the aggregation module.
    Improvements of up to five VQA datasets from the experiments conducted are evident.
    Following the results, this study analyses the effects of both the improvements on the domain generalization problem.
    The code is available on GitHub though the following link \url{https://github.com/bhanukaManesha/ramen}.
\end{abstract}


\section{Introduction}

Visual Question Answering (VQA) is a multi-disciplinary problem in machine learning that exists at the intersection of the computer vision, natural language processing and knowledge representation fields \cite{Antol2015}.
Recently, the task of VQA has been classified as an AI-complete task due to the complexity of it.
This problem requires the semantic understanding of each of the three fields as well as the relationship between each one of them \cite{fusionsurvey}.
One of the main issues in VQA is that the state-of-the-art (SOTA) results on the datasets do not translate on to real-world applications. 
This has directed the VQA field towards generalization. 

The datasets in the field of VQA can be separated into two main categories \cite{Shrestha2019}.
The first type focuses on answering questions by understanding the objects on natural real world images and the other focuses on using synthetic images to test reasoning questions.
The problem with this categorization is that the algorithms tend to focus on one or the other and not generalize on both.
This known as the domain generalization problem, because the VQA models generalize on both types of dataset either through training from scratch or fine tuning to the domains and not overfitting on one type.
\cite{Shrestha2019} addressed this issue by introducing a framework for domain generalization.
This framework allows to train models of both domains with similar visual and textual features to evaluate their generalization ability.

They also introduced the RAMEN model architecture which was able to outperform all the other models compared in the study in terms of domain generalization.
However, this model uses a simple architecture with a potential for improvement and exploration.
Therefore, this study proposes improvements to the architecture of the RAMEN model while analyzing the effect of these changes to the overall problem of domain generalization.

The main contributions of this study includes the following:
\begin{itemize}
    \item Improvements to the domain generalization performance of the RAMEN model architecture by proposing modifications to the fusion and aggregation modules.
    \item A broad comparison of the vector based fusion operations for early and late fusion pertaining to domain generalization.
    \item Implementation and analysis of a transformer based aggregation module to map the relationships between bi-modal embeddings of the regional proposals in the RAMEN model.
\end{itemize}

The rest of the paper is organized as follows:
Section \ref{relatedwork} provides more context to the domain generalization problem in VQA, the RAMEN model and the transformer architecture in VQA.
The proposed improvements to the RAMEN model is detailed in Section \ref{methodology}, which is followed by the experiment strategy used in Section \ref{experiment}.
A comprehensive analysis of the results is conducted in Section \ref{result} which is then summarized in Section \ref{conclusion}.

\section{Related Work} \label{relatedwork}
\begin{figure*}
    \centering
    \includegraphics[width=15cm]{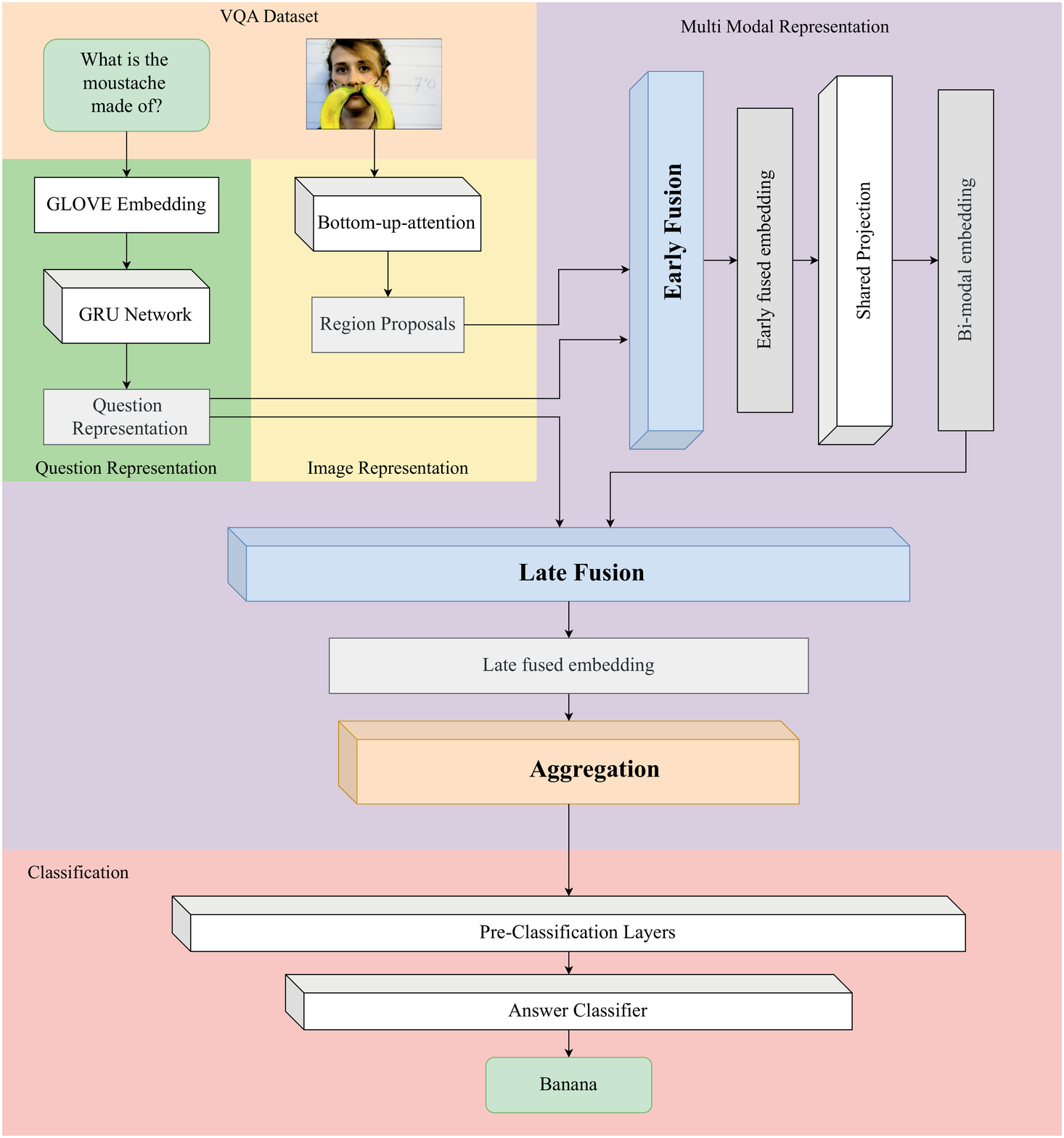}
    \caption{High-level architecture of the RAMEN model}
    \label{fig:methodology}
  \end{figure*}
  
\begin{table*}[t]
    \centering
    \caption[Summary of VQA Datasets]{Summary of VQA datasets used in this study. \cite{Shrestha2019}}
    \label{dataset-summary} 
        \begin{threeparttable}
        \begin{tabular}{|J|L|L|L|L|Y|}
            \hline
            \textbf{Dataset} & \textbf{Image Type} & \textbf{Question Type} & \textbf{Images} & \textbf{Q\&A Pairs} & \textbf{Download links} \\ \hline
            VQAv1 \cite{Antol2015}& Natural & Human & 204K & 614K & \url{https://visualqa.org/vqa_v1_download.html} \\ \hline 
            VQAv2 \cite{Goyal2019}& Natural & Human & 204K & 1.1M & \url{https://visualqa.org/download.html}\\ \hline
            TDIUC \cite{Kafle2017} & Natural & Both & 167K & 1.6M & \url{https://kushalkafle.com/projects/tdiuc.html}\\ \hline
            C-VQA  \cite{Agrawal2017}& Natural & Human & 123K & 369K & \url{https://computing.ece.vt.edu/~aish/cvqa/}\\ \hline
            VQACPv2 \cite{agrawal2018don} & Natural & Human & 219K & 603K & \url{https://computing.ece.vt.edu/~aish/vqacp/}\\ \hline
            CLEVR \cite{Johnson20171}& Synthetic & Synthetic & 100K & 999K & \url{https://cs.stanford.edu/people/jcjohns/clevr/}\\ \hline
            CLEVR-Humans \cite{Johnson20172}& Synthetic & Human & 32K & 32K & \url{https://cs.stanford.edu/people/jcjohns/iep/}\\\hline
            CLEVR-CoGenT-A \cite{Johnson20171}& Synthetic & Synthetic & 100K & 999K & \url{https://cs.stanford.edu/people/jcjohns/clevr/}\\ \hline
            CLEVR-CoGenT-B \cite{Johnson20171}& Synthetic & Synthetic & 30K & 299K & \url{https://cs.stanford.edu/people/jcjohns/clevr/}\\ \hline
        \end{tabular}
    \end{threeparttable}
\end{table*}

This section first summarizes the main VQA datasets used in this study, followed by the RAMEN model.
Next, the background of the transformer architecture in VQA is explored.

\subsection{VQA Datasets}
The dataset is the most important part of the VQA pipeline as it determines what the model learns.
If the dataset contains inherent biases, the model will learn these and the performance of the model will be affected.
Many datasets were introduced, with each dataset focusing on either solving biases \cite{malinowski2014nips, Agrawal2017, agrawal2018don, Johnson20171, krishnavisualgenome} or solving a specific type of question domain \cite{Malinowski2018, Hasan2018, Shah2019, Kafle2018}.

This study focuses on the datasets used by \cite{Shrestha2019}, where the datasets were divided into two groups; VQA datasets for natural image understanding and VQA datasets to test reasoning.
Table \ref{dataset-summary} summarizes the VQA datasets used by \cite{Shrestha2019} to test for generalization.

\subsubsection{\textbf{VQA Datasets for Natural Image Understanding}}
These datasets aims to provide answers by identifying objects in the image.
This can be through colour, count or other visual cues.
All the datasets in this group uses the MSCOCO dataset \cite{cocodataset} as the base image dataset except for TDIUC which adds extra images.

\paragraph{VQAv1 \cite{Antol2015}}
One of the most widely known datasets with the current SOTA accuracy of 75.26\%.
This dataset mainly focuses on detection questions such as \emph{Is there food on the table} \cite{Antol2015} and lacks reasoning questions such as \emph{What is behind the computer in the corner of the table?} \cite{malinowski2014nips}.
However, it consists of inherent question-answer biases where questions such as \emph{Where is the giraffe standing on?} always contains the answer \emph{grass}.

\paragraph{VQAv2 \cite{Goyal2019}}
The successor to VQAv1, was able to reduce the question-answer biases by introducing complementary questions.
However, even though this allowed the VQAv2 dataset to be more balanced, the bias of having more detection questions is still prevalent in this dataset; which makes the models trained on VQAv2 datasets inherently weaker when answering questions with reasoning.

\paragraph{TDIUC \cite{Kafle2017}}
This dataset was created with the primary aim of evaluating the performance of models on 12 distinct types of VQA tasks. 
Color attributes, positional reasoning and object presence are some of the types of tasks.
A new metric called Mean-per-type was also introduced as shown in Equation \ref{tdiuc_metrics} in Section \ref{tdiuc_metrics}.
Therefore, it is evident that a model needs to perform well across all the question types to get a good performance score.

\paragraph{C-VQA \cite{Agrawal2017}}
Aims to re-split the VQAv1 dataset to introduce novel combinations for the question-answer pairs when testing.
During testing the models will come across new combinations of question-answer pairs.
Therefore, the models need to be able to generalize on the task and not the question and answer.

\paragraph{VQACPv2 \cite{agrawal2018don}}
Overcomes the question and language bias by splitting the VQAv1 and VQAv2 dataset.
A completely different answer distribution is present in the test split compared to the training split.
This allows the models to test their ability to generalize by not over-fitting on the training set.

\subsubsection{\textbf{VQA Datasets to Test Reasoning}}

These datasets aim to test the ability of models to answer reasoning based questions by using synthetic images.
These synthetic computer generated images allow this dataset to generate complex reasoning questions automatically.
All the datasets in this group use the images from the CLEVR dataset, with each dataset having different question-answer pairs.

\paragraph{CLEVR \cite{Johnson20171}}
The main goal of this dataset is to test the reasoning capability of models on geometric shapes.
Similar to TDIUC, this dataset is classified into five categories.

\paragraph{CLEVR-Humans \cite{Johnson20172}}
The main downside of CLEVR dataset is that the questions are computer generated, thus being very structured.
The CLEVR-Humans dataset addresses this issue by using free form human generated question-answer pairs.
It still uses the same images from the CLEVR dataset.

\paragraph{CLEVR-CoGenT \cite{Johnson20171}}
This dataset was introduced with the CLEVR dataset having two splits with mutually exclusive color and shapes, namely, CLEVR-CoGenTA and CLEVR-CoGenTB.
This dataset aims to study the model’s ability to recognize novel combinations of attributes such as color and shapes at test time. 
For example, CLEVR-CoGenTA contains red colour cylinders in the training set, in contrast, CLEVR-CoGenTB does not contain red colour cylinders.

\subsection{RAMEN}

The VQA pipeline consists of five main components; VQA dataset, Image representation, Question representation, Multi-modal representation and Answer classification.
Many studies have been done in the field of VQA, with each focusing on improving different sections of the VQA pipeline \cite{Antol2015, Xu2016, lu2016hierarchical, Ilievski2016,  Noh2016, Yang2016, Nam2016, fukui2016multimodal, Anderson2018, Jiang2018,Kim2018, Liu2019, Yu2019}.

\cite{Shrestha2019} proposed a framework to compare the performance of VQA algorithms across different domains.
They standardize the image representation and the question representation across the VQA datasets of multiple domains.
With this, they were able to compare the performance of multiple algorithms \cite{Anderson2018, Norcliffe-brown2018, Kim2018, Hudson2018, Malinowski2018} across domains and assess the generalization ability of the model architectures.

They also proposed a model named RAMEN with a conceptually simple architecture that was able to generalize across multiple domains.
Figure \ref{fig:methodology} shows the high-level architecture of the RAMEN model, where the five main components of the VQA pipeline can be identified in this model.

\paragraph{Image Representation}

The image representation module focuses on extracting features from the image and converting them into visual features.
Various methods exists that focuses on extracting features from images using different techniques such as VGG-Net \cite{Antol2015, chen2015abc, lu2016hierarchical, xiong2016dynamic, ren2015exploring}, ResNet \cite{Nam2016, ben2017mutan, fukui2016multimodal, Ilievski2016, su2018learning} and Faster-RCNN \cite{Anderson2018, singh2018attention, song2018pixels, lu2017co}.
The RAMEN model uses a Faster-RCNN based technique where the image is passed through the bottom-up-top-down network \cite{Anderson2018}, which uses attention on the object level to return visual features as a set of regions.
These regions corresponds to the main object regions in the image which are used to answer the questions.
For VQAv1, VQAv2, CVQA, VQACPv2 and TDIUC datasets the bottom-up attention module returns 36 regions and for the CLEVR family of datasets it return 15 regions.

\paragraph{Question Representation}

The question representation module converts the question into a vector representation.
This vector representation encodes all the words while maintaining the flow and positional information of the question.
Studies have proposed multiple ways to extract these features by using CNN \cite{Yang2016, ma2015learning}, LSTM \cite{fukui2016multimodal,chen2015abc, Ilievski2016} and GRU \cite{Anderson2018, ben2017mutan, song2018pixels} networks.
The RAMEN model uses a GRU based approach by first splitting the questions into multiple word tokens.
Each token is instantiated with the GLOVE embedding \cite{glove}.
Then the embeddings are passed through a GRU based RNN \cite{gru} to obtain the question representation.

\paragraph{Multi Modal representation}

Once the image and question representations are passed into this module, the two vectors are fused together using concatenation.
This step is also known as early fusion in the model architecture.
Next, the RAMEN model uses a Multi Layer Perceptorn (MLP) to create a bimodal embedding.
This allows the model to learn the relationship between the image and question representation.
Then the bimodal embedding is concatenated with the question representation, with the Late Fusion module.
The fused vector is then passed to the aggregation module where it is passed through a bi-directional GRU network.
This step captures the relationships between the bimodal embeddings.

\paragraph{Classification}
In the final module, the output of the multi-modal representation is pass through a series of linear layers to perform the pre-classification step.
This is then followed by a single linear layer for classification.

\subsection{Transformer}

The introduction of the transformer architecture \cite{vaswani2017attention} has been a pivotal moment in the NLP community.
The main use case of the transformer network is for machine translation.
The main advantage of using transformer over traditional RNN networks is that the sequences are processed as a whole compared to one by one.
Moreover, the transformer uses multi-head attention and positional encoding to obtain more information about the relationships between the features.
This allows the transformer architecture to be parallelizable compared to sequential RNN networks.

Many studies have been done in the field of VQA that incorporated transformers into the architecture \cite{li2019entangled, sur2020self, kant2020spatially}.
Most of them focus on encoding the question using the transformer architecture.
The Bidirectional Encoder Representations from Transformers (BERT) architecture \cite{su2019vl}, which is derived from the transformer architecture is commonly used to encode the question \cite{tan2019lxmert, khan2020mmft}.
However, limited research has been done where the transformer architecture is used to capture the relationship between visual and question features.
\cite{tan2019lxmert} showed that the transformer architecture is able to capture intra-modality and cross-modality relationships on the VQA and GQA \cite{hudson2019gqa} datasets.
Therefore, this study aims to investigate the effect of using transformer as an encoder in VQA.

\section{Methodology} \label{methodology}

\begin{figure*}[t]
  \centering
  \begin{subfigure}{0.35\textwidth}
    \centering
    \includegraphics[width=\textwidth]{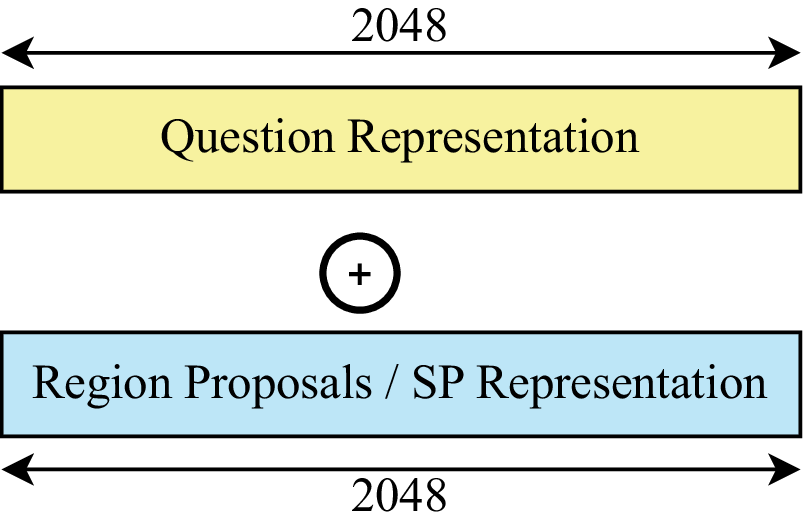}  
    \caption{Additive Fusion}
    \label{fig:addfusion}
  \end{subfigure}
  \hspace{1.5em}
  \begin{subfigure}{0.35\textwidth}
    \centering
    \includegraphics[width=\textwidth]{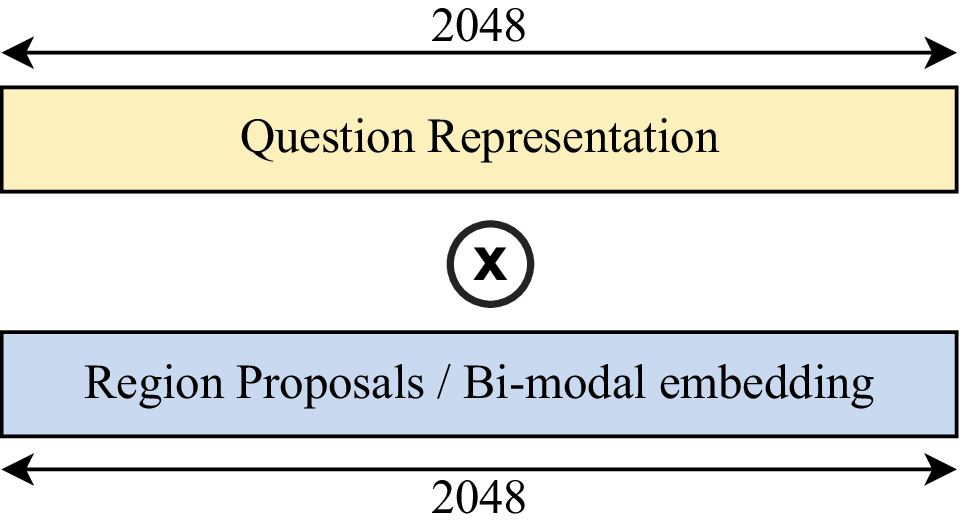}  
    \caption{Multiplicative Fusion}
    \label{fig:multfusion}
  \end{subfigure}
  \par\bigskip
  \begin{subfigure}{0.7\textwidth}
      \centering
      \includegraphics[width=11cm]{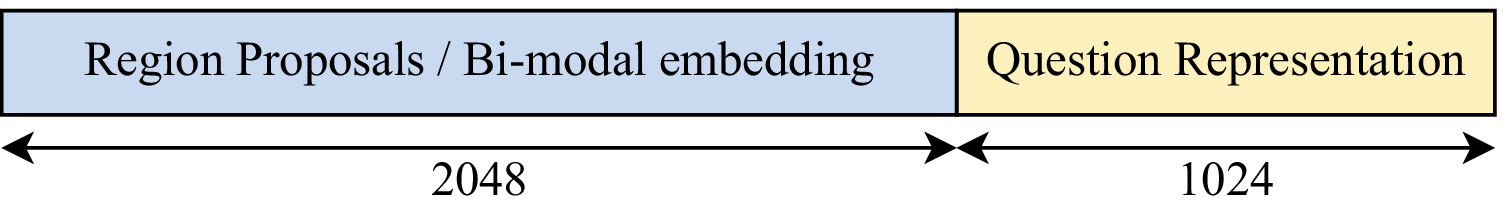}  
      \caption{Concat Fusion}
      \label{fig:concatfusion}
    \end{subfigure}

    \par\bigskip
    \begin{subfigure}{0.7\textwidth}
      \centering
      \includegraphics[width=12cm]{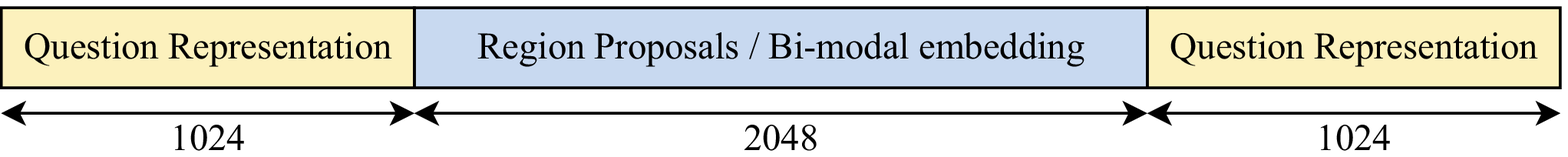}  
      \caption{Question Fusion}
      \label{fig:questionfusion}
    \end{subfigure}

  \caption{Fusion Strategies for Early and Late fusion}
  \label{fig:fusions}
  \end{figure*}

The key focus of the improvements are on the multi modal representation section of the RAMEN model.
Based on the ablation study done by \cite{Shrestha2019}, the early fusion module has a significant effect on the performance of the model.
The aggregation technique also has an effect on the performance of the model, whereas, the late fusion module has the minimum effect on the performance of the model.
Therefore, the experiments are done on the Early Fusion, Late Fusion and the Aggregation modules of the model as shown in Figure \ref{fig:methodology}.

\subsection{Fusion Strategies}
\cite{fusionsurvey} performed a survey on the fusion strategies in ImageVQA and VideoVQA studies.
They classified the fusion strategies into three main types; Vector operations, Neural Networks (NN) and Bilinear pooling.

The RAMEN model uses a mix of vector operations and neural networks to perform the multi-modal fusion.
The early and late sub-modules uses simple concatenation of the features and the shared projection and aggregation sub-modules uses neural networks as the strategy.
First in the early fusion module, the regional visual features are fused using concatenation with the question embedding to obtain the early fused embedding.
This is then passed through the neural network based shared projection and the output bi-modal embedding is obtained.
In the late fusion module, the bi-modal embedding is again fused using concatenation with the question embedding to obtain the late fused embedding.
After the fusion operation, both early and late fusion embeddings are passed through a Batch Normalization \cite{ioffe2015batch}.
Finally, the vector is passed through the aggregation module, which is a Recurrent Neural Network based fusion strategy to obtain the fused vector for classification.

In the survey, \cite{fusionsurvey} also categorized the vector operations into three main sections; concatenation, addition and multiplication.
In this study, these three main vector operations will be experimented on the RAMEN model to observe the performance effect on the NN.
Figure \ref{fig:fusions} shows the overview of the fusion strategies tested in this study.

\subsubsection{\textbf{Concat Fusion}}
This is the baseline strategy used by the RAMEN model.
A question embedding size of 1024 and visual feature size of 2048 is used to obtain a final embedding size of 3072.
In this approach the output embedding passes all the information from both embeddings to the NN to identify the relationships.
No information is lost in this approach and all the feature points are given similar weights.
In order to perform vector operations, the question embedding is repeated to match the size of the visual features and bi-modal embedding.
To do this, the question embedding is repeated 36 times for the VQA family of datasets and 15 times for the CLEVR family of datasets.
This is done for all the fusion strategies experimented in this study.

Equation \ref{eq:concatanation} is used to obtain the final embedding ($c_i$), where $q_i$ is the question embedding and $v_i$ is the regional visual features or the bi-modal embedding.

\begin{equation}
  c_i = BatchNorm([q_i , v_i])
  \label{eq:concatanation}
 \end{equation}

\subsubsection{\textbf{Additive Fusion}}
For the additive fusion, the question embedding is matched to the same size as the visual features.
Therefore, the embedding size is changed from 1024 to 2048 to obtain the final embedding size of 2048.
This approach emphasizes on the different feature points which allows the model to update the question embeddings to focus on them.
This approach has information loss due to the addition operation, however, it is compensated by the increase in the question embedding size.
Equation \ref{eq:additive} is used to obtain the final embedding ($c_i$).
$q_i$ and $v_i$ are as described as in Equation \ref{eq:concatanation}.

\begin{equation}
  c_i = BatchNorm(v_i \oplus q_i)
  \label{eq:additive}
 \end{equation}

\subsubsection{\textbf{Multiplicative Fusion}}
Similar to additive, the question embedding size of 2048 is used for this strategy.
In this approach, the emphasis on different feature points is greater than the additive fusion.
Equation \ref{eq:multiplicative} is used to obtain the final embedding ($c_i$).
$q_i$ and $v_i$ are as described as in Equation \ref{eq:concatanation}.

\begin{equation}
  c_i = BatchNorm(q_i \odot v_i)
  \label{eq:multiplicative}
 \end{equation}

\subsubsection{\textbf{Question Fusion}}
Question fusion uses a dual concatenation strategy.
The question embedding is concatenated before and after the visual features.
An embedding size of 1024 is used for the question embedding, with the final embedding size of 4096.
The main emphasis in this strategy is to provide more feature points for the question embedding.
Datasets such as the CLEVR family that are used to test reasoning contains longer questions compared to other datasets.
Therefore, limiting the question embedding to a single vector of size 1024 or 2048 can effect the emphasis of the question on the model.
Equation \ref{eq:questionconcatanation} is used to obtain the final embedding ($c_i$).
$q_i$ and $v_i$ are as described as in Equation \ref{eq:concatanation}.

\begin{equation}
  c_i = BatchNorm([q_i , v_i , q_i])
  \label{eq:questionconcatanation}
 \end{equation}

\subsection{Aggregation Strategies}

This module is used to calculate the relationship between the question and bi-modal embeddings.
The bi-modal embeddings contains the relationship between the question and each regional visual feature.
Thereby this module aims to identify the relationships between the visual regions.
Higher performance on this module will lead to better results on questions that require multi object or localized information to answer.

\subsubsection{\textbf{bi-GRU network}}
The baseline aggregation strategy in the RAMEN model uses a bidirectional GRU based RNN to calculate the feature vector.
The main downside to this approach is that the model goes through each region sequentially in both directions.
Therefore, to obtain a relationship between two regions of the image, the model needs to pass through the other regions which can lead to information loss.
This approach is best used when all the regions are equally important for the question.

\subsubsection{\textbf{Transformer network}}
The transformer architecture is stronger in identifying the relationship between the multiple regions/vectors, because the network processes all the regions at once and not sequentially.
This is the reason why the transformer model performs well on the machine translation tasks \cite{singh2018attention}.
This allows it to capture relationships among regions better than RNNs.

However, the positional encoder in the traditional transformer network masks out half of the regions.
This is to ensure the model is not able to see the next word in machine translation.
For the RAMEN model, this is not an issue. 
So the mask is removed and the transformer is able to view all the regions.

The output of the original transformer model is a set of decoders for the translated sentence.
But in this case, the main aim is to obtain a representation to be passed to the classification module.
Therefore, the decoder is replaced with a fully connected NN that returns a vector representation instead of the transformer decoder module.

One of the main downside of the transformer network is the slow convergence.
Typically the transformer network might take upto sixty hours to fully convergence on translation tasks.

\section{Experiment} \label{experiment}

\subsection{Dataset specification}

In the baseline paper, the accuracy of the CLEVR-CoGenTB dataset was obtained on a sub split of the test set.
But in this study, the accuracy is obtained on the complete test set.
Similarly, the original paper fine-tuned the model trained on the CLEVR-dataset with the CLEVR-Humans dataset to obtain the accuracy.
However, this study, trains the CLEVR-Humans dataset from scratch.
All other training and testing splits of the datasets are identical to the baseline paper.

\subsection{Model specification}
Due to the changes in the datasets, the baseline accuracies are all re-calculated to ensure consistency.
All the model hyper-parameters are maintained as mentioned in the baseline paper.

The model with the transformer as the aggregation strategy is named as the TransformerNet and for baseline model with the bi-GRU network the name RAMEN model is used.
With each model, the four different fusion strategies are experimented for both the early and late fusion modules.
Therefore, in total the nine datasets are trained on eight versions of the models.

\subsection{Evaluation metrics}

Three types of evaluations metrics are used in this study to compare the results between the datasets.
These are the same metrics used in the baseline study.

\paragraph{10-choose-3}

Equation \ref{vqa_metric} shows the evaluation metric used by VQAv1, VQAv2, CVQA and VQACPv2.
These datasets provide multiple answers for each question from multiple human annotators.
Thus using this metric reduces the inter-human variability \cite{Antol2015}.

\begin{equation}\label{vqa_metric}
	Acc(answer) = min\{ \frac{\text{\# of annotators provided answer}}{3}, 1 \}
\end{equation}

\paragraph{Simple Accuracy}

CLEVR, CLEVR-Humans, CLEVR-CoGenT-A and CLEVR-CoGenT-B uses the simple accuracy shown in Equation \ref{clevr_metric} as the evaluation metric \cite{Johnson20171}. 

\begin{equation}\label{clevr_metric}
	Acc(answer) = \frac{\text{\# correct answer}}{\text{\# questions}}
\end{equation}

\paragraph{Mean-per-type} \label{sec:mean-per-type}

The TDIUC dataset uses the mean-per-type evaluation metric as shown in Equation \ref{tdiuc_metrics} \cite{Kafle2017}.
This ensures that the model is able to perform well on each category, even though the number of test instances of each category are different. 

\begin{equation}\label{tdiuc_metrics}
	Acc(answer) = \frac{\sum{\{\frac{\text{\# correct answer per type}}{\text{\# of questions per type}}}\}}{\text{\# of types}}
\end{equation}

\subsection{Training specifications}

All the experiments were done on a PC running \emph{Ubuntu 18.04.1 LTS} with an \emph{Intel® Xeon(R) W-2145 CPU @ 3.70GHz} with 16 logical cores and 64GB RAM.
A single \emph{Quadro P5000} GPU was used to perform the NN training with a 7200RPM \emph{Seagate} hard drive to store the data.
The gradual learning rate warm up is used similar to \cite{Kim2018,Shrestha2019}.
The mini-batch size of 256 is used for all the experiments.
The models are trained until 25 epochs with some exceptions in the TransformerNet experiments; mainly due to the slower convergence rate.
As shown in Appendix \ref{apdx:trainingtime}, an average training time of 46 minutes per epoch was elapsed for all experiments.

\section{Results \& Discussion} \label{result}

\begin{table*}
    \centering
    \caption{Results from all eight model with the nine VQA datasets.}
    \arrayrulecolor{black}
    \begin{tabular}{l|c|ccc|cccc}
        \multirow{2}{*}{\textbf{Dataset }} & \multicolumn{4}{c}{\textbf{Ramen }}                                                                                                                                                                      & \multicolumn{4}{c}{\textbf{TransformerNet }}                                                                 \\
                                           & \begin{tabular}[c]{@{}c@{}}\textbf{Baseline}\\\textbf{Concat} \cite{Shrestha2019} \end{tabular} & \textbf{Additive}                         & \textbf{Multiplicative}                            & \textbf{Question}                                  & \textbf{Concat}                           & \textbf{Additive} & \textbf{Multiplicative} & \textbf{Question}  \\
        \hline
                                           \textbf{VQAv1}                     & 63.30                                              & 63.21                                     & {\cellcolor[rgb]{0.455,0.655,0.996}}\textbf{65.54} & {\cellcolor[rgb]{0.659,0.776,0.996}}64.76          & {\cellcolor[rgb]{0.831,0.89,0.996}}63.32  & 55.88             & 60.91                   & 55.08              \\
        \textbf{VQAv2}                     & 62.16                                              & 63.64                                     & {\cellcolor[rgb]{0.455,0.655,0.996}}\textbf{65.28} & {\cellcolor[rgb]{0.659,0.776,0.996}}65.07          & {\cellcolor[rgb]{0.831,0.89,0.996}}64.06  & 59.79             & 56.32                   & 50.14              \\
        \textbf{VQACPv2}                   & {\cellcolor[rgb]{0.455,0.655,0.996}}\textbf{37.61} & 36.73                                     & 36.28                                              & {\cellcolor[rgb]{0.831,0.89,0.996}}37.03           & {\cellcolor[rgb]{0.659,0.776,0.996}}37.47 & 27.60             & 27.60                   & 26.89              \\
        \textbf{CVQA}                      & {\cellcolor[rgb]{0.659,0.776,0.996}}56.98          & 55.82                                     & 56.58                                              & {\cellcolor[rgb]{0.831,0.89,0.996}}56.81           & {\cellcolor[rgb]{0.455,0.655,0.996}}\textbf{57.74} & 53.49             & 54.20                   & 48.47              \\
        \textbf{TDIUC}                     & {\cellcolor[rgb]{0.455,0.655,0.996}}\textbf{66.48} & 64.90                                     & {\cellcolor[rgb]{0.659,0.776,0.996}}65.69          & 64.81                                              & {\cellcolor[rgb]{0.831,0.89,0.996}}65.47  & 58.03             & 56.34                   & 53.90              \\
        \hline
        \textbf{CLEVR}                     & {\cellcolor[rgb]{0.659,0.776,0.996}}96.52          & {\cellcolor[rgb]{0.831,0.89,0.996}}96.26  & {\cellcolor[rgb]{0.455,0.655,0.996}}\textbf{96.72} & 96.06                                              & 95.79                                     & 50.52             & 57.31                   & 50.07              \\
        \textbf{CLEVR-Humans}              & 44.57                                              & 40.21                                     & {\cellcolor[rgb]{0.831,0.89,0.996}}46.46           & {\cellcolor[rgb]{0.455,0.655,0.996}}\textbf{48.63} & {\cellcolor[rgb]{0.659,0.776,0.996}}46.49 & 38.45             & 40.07                   & 37.99              \\
        \textbf{CLEVR-CoGenTA}             & 96.59                                              & {\cellcolor[rgb]{0.659,0.776,0.996}}96.84 & {\cellcolor[rgb]{0.831,0.89,0.996}}96.63           & {\cellcolor[rgb]{0.455,0.655,0.996}}\textbf{96.90} & 96.43                                     & 64.26             & 72.07                   & 60.24              \\
        \textbf{CLEVR-CoGenTB}             & 88.27                                              & {\cellcolor[rgb]{0.659,0.776,0.996}}89.42 & 86.22                                              & {\cellcolor[rgb]{0.831,0.89,0.996}}88.74           & {\cellcolor[rgb]{0.455,0.655,0.996}}\textbf{89.68} & 55.80             & 60.18                   & 55.19              \\
        \hline
        \textbf{Mean}                      & 68.05                                              & 67.45                                     & {\cellcolor[rgb]{0.831,0.89,0.996}}68.38           & {\cellcolor[rgb]{0.455,0.655,0.996}}\textbf{68.76} & {\cellcolor[rgb]{0.659,0.776,0.996}}68.49 & 51.53             & 53.89                   & 48.66             
        \end{tabular}
    \arrayrulecolor{black}
    \label{tab:results}
    \end{table*}

    \begin{figure*}[ht]
        \centering
        \begin{subfigure}{0.47\textwidth}
            \centering
            \includegraphics[width=\linewidth]{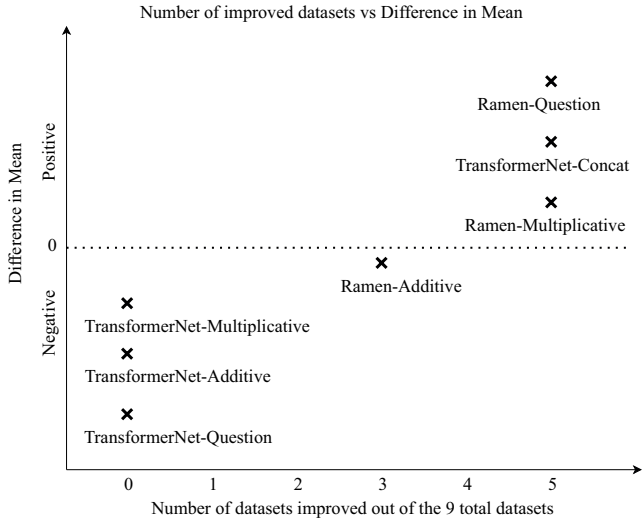}
            \caption{The difference in mean compared to the overall number of improved datasets.}
            \label{fig:improvement1}
        \end{subfigure}
        \hfill
        \begin{subfigure}{0.47\textwidth}
            \centering
            \includegraphics[width=\linewidth]{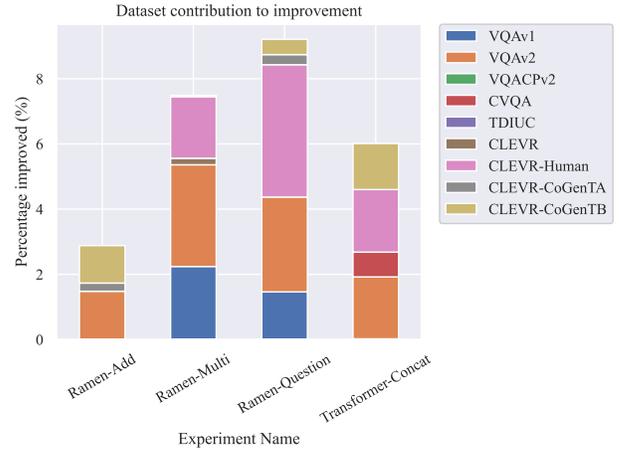}
            \caption{Improvement of the datasets for each model compared to the baseline.}
            \label{fig:improvement2}
        \end{subfigure}
        \par\bigskip
        \caption{Summaries of the results.}
        \label{fig:improvements}
    \end{figure*}
    
This section focuses on the key observations from the experiments and the effect of changing the aggregation and fusion strategies.
Table \ref{tab:results} show the scores obtained by each experiment.
The top three models are highlighted with darker colours indicating better performance. 
Appendix \ref{apdx:results} contains the full results table with the training scores and number of epochs.

When comparing the baseline results (\emph{Ramen-Concat}) with the results from the RAMEN study \cite{Shrestha2019}, it is evident that there exist a minor difference in the scores.
The training was done with the same hyper-parameters as stated in the baseline paper even though there exists a deficiency in the scores.
However, since all the experiments are done with the fixed set of hyper-parameters, the scores in this study are consistent.
      
\subsection{Overall observations} \label{res:overall}
First, considering the model with the highest mean score across all the datasets, the \emph{Ramen-Question} model has a score of 68.76.
With a percentage difference of about 1\%, it is evident that the improvement of using different fusion and aggregation strategies is minor.
However, many other patterns in the results can be observed which can help improve the performance of future models.
It is also noted that \emph{Ramen-Multiplicative} and \emph{TransformerNet-Concat} was also able to improve the performance by about 0.5\% and 0.65\% respectively.

Next, it is evident that TransformerNet model did not perform at all on most of the datasets, with differences in accuracies of more than 25\% on the CLEVR datasets.
This issue is addressed in Section \ref{sec:aggregation}.
However, the \emph{TransformerNet-Concat} model performed well on most of the datasets.

When considering the model with the highest number of top scores on the nine datasets, the \emph{Ramen-Multiplicative} model has achieved the highest score on three of the main datasets.
Therefore, it is evident that this model can perform well on both natural and synthetic types of VQA datasets.
However, the model is not able to generalize to question and attribute biases well.
This is identified from the CVQA and VQACPv2 datasets, due to the lower performance on them.
When comparing the performance of the model between CLEVR-CoGenTA and CLEVR-CoGenTB datasets, it is found to be evident that the model is seeing a dip in performance.

Next, comparing the performance of the model based on the scores in the top-3 rank of the experiments, it is evident that \emph{Ramen-Question} has the overall best performance.
In seven out of the nine dataset, this model was able to achieve the top three results.
This indicates the ability of the model to generalize across multiple datasets.
Also the \emph{TransformerNet-Concat} model was able to achieve top three in seven out of nine datasets even though it only have the highest scores for the CVQA dataset.

With reference to Figure \ref{fig:improvement1}, the \emph{Ramen-Additive} model was able to improve on three datasets.
However, due to the lower scores on the other datasets, especially CLEVR-Humans, the model is not able to achieve a positive mean.
This indicates that the model is not able to perform well on free-form questions.

In the following section the performance of the models on specific datasets are analysed.

\subsection{Dataset observations}

Figure \ref{fig:improvement2} demonstrates the datasets impact on the improvement of the models.
It only showcase the experiments with at least one improvement, hence \emph{TransformerNet-Additive}, \emph{TransformerNet-Multiplicative} and \emph{TransformerNet-Question} are ignored.
All of the remaining models were able to improve on the VQAv2 dataset.
This suggests that all the fusion strategies and the transformer aggregation strategy can improve the performance of localizing and detecting items.

However, only the Multiplicative and Question fusion were able to improve the VQAv1 dataset.
Since, VQAv1 has inherent bias in the questions, this implies that the these two fusion strategies are more prone to over-fitting on the VQA based datasets.
This is also made clear by the poor performance on the CVQA and VQACPv2 datasets.

However, the most performance gain is by the \emph{Ramen-Question} on the CLEVR-Humans dataset.
It points to an indication that the pre and post concatenation of the question embedding has an improvement on the free form questions.
This is also true for the models performance on VQAv1 and VQAv2 datasets.

Compared to the baseline, the \emph{TransformerNet-Concat} is the only model to have an improvement in the score for CVQA.
This highlights that the transformer aggregation module provides the ability for the module to generalize.
This is further emphasized since the CLEVR-CoGenTB and VQAv2 datasets also show improvement in the scores.

\subsection{Fusion Strategies} \label{res:fusionstrat}
Overall it is clear that the different fusion strategies favor various datasets due to the unique characteristics in them.

\subsubsection{Concat Fusion}

The baseline concatenation fusion approach was able to get the highest score for VQACPv2 and TDIUC dataset using the RAMEN model and CLEVR-CoGenTB using the TransformerNet model.
The VQACPv2 dataset aims to test the answer bias in the models.
Therefore, concatenation based fusion is able to generalize well in terms of answer biases as both \emph{Ramen-concat} and \emph{TransformerNet-Concat} was able to achieve high scores.

Next, the TDIUC MPT metric measures the performance of the model on multiple question types.
Considering that both the \emph{Ramen-Concat} and \emph{TransformerNet-Concat} have high scores on the TDIUC dataset, it is clear that concatenation based fusion allows for much more question type based generalization.

The \emph{TransformerNet-Concat} model is trained on CLEVR-CoGenTA and tested on CLEVR-CoGenTB.
Therefore, the model will not learn any details about the complementary attributes in the dataset.
This indicates that the model is able to generalization well onto unseen combinations of attributes.
However, the relationship between concatenation based fusion and attribute based generalization cannot be established.
This is due to the lower score in the \emph{Ramen-Concat} model, which implies that the performance gain is due to the transformer aggregation strategy.

\subsubsection{Additive Fusion}

The additive fusion strategy was not able to get the highest score for any of the datasets.
However, the \emph{Ramen-Additive} model was able to improve on the CLEVR-CoGenTA and CLEVR-CoGenB datasets, which point towards the model's ability to generalize to new concept compositions.
The main issue with the additive fusion strategy is the information loss and the lower emphasis on the vector operation.

\subsubsection{Multiplicative Fusion}

The multiplicative fusion strategy achieved the highest score for VQAv1, VQAv2 and CLEVR datasets.
As mentioned in Section \ref{res:overall}, the model suffers with generalization.
However, the emphasis on the vector operation is higher compared to additive fusion, therefore the most important details are passed on through the fusion module.

\subsubsection{Question Fusion}

The question fusion obtained the highest score for CLEVR-Humans and CLEVR-CoGenTA.
It also achieved high scores for all the CLEVR datasets.
This is an indication that for reasoning type datasets with higher significance on the question, the double concatenation of the question has an effect.

The ability for the model to generalize on new concept compositions can be observed due to the performance on CLEVR-CoGenTA and CLEVR-CoGenTB.

\subsection{Aggregation Strategies} \label{sec:aggregation}
The transformer module as the aggregation strategy does not perform well.
With only decent performance using the concat fusion strategy, it may not be suitable to be part of the RAMEN model.
Many issues where faced when training the transformer module such as slow convergence and longer training time.
However, the slow convergence of the transformer module remains a significant drawback.

This is evident when considering the number of epochs used to train \emph{TransformerNet-Concat} vs \emph{TransformerNet-Question} on the VQAv2 dataset.
\emph{TransformerNet-Concat} was trained for 50 epochs where the highest score was obtained at epoch 46, whereas \emph{TransformerNet-additive} was only trained for 25 epochs.
Appendix \ref{apdx:results} reports all the training details used for each dataset.
Therefore, training the TransformerNet models for longer can provide better scores.

Additionally, due to the longer training times and the time constraint, hyper-parameter tunning was not an option.
With an average time of 58 minutes per epoch for the \emph{TransformerNet-Question} model, the training would take more than 48 hours on a single GPU.
However, as observed by \emph{TransformerNet-Concat}, in an ideal scenario the model is able to convergence.

\section{Conclusion} \label{conclusion}
The proposed improvements of this study resulted in minor gains in performance of about 1\%.
However, in-terms of domain generalization the \emph{Ramen-Multiplicative}, \emph{Ramen-Question} and \emph{TransformerNetwork-Concat} models were able to achieve improvements in five out of the nine datasets.
Also, \emph{Ramen-Question} and \emph{TransformerNetwork-Concat} models were able achieve the top three scores in seven out of the nine datasets.

Analyzing the fusion strategies, provided insights to the different characteristics that effect domain generalization.
For example, Question fusion performed well on reasoning questions due to the increase in the number of question embedding data points, which resulted in an increase in the amount of information passed into the aggregation module.
This knowledge can be used to improve the performance of model and domain generalization.

When studying the effects of the Transformer module as the aggregation strategy, it is clear that selecting the correct hyper-parameters and providing the necessary amount of training time to converge are two main requirements when training the transformer module.
This is one of the main limitations of this study.
The time constraint and high computational cost of the experiments led to some of the experiments not converging to the higher scores.
Therefore, since VQA datasets tend to be larger in size, more powerful hardware is needed to perform a broader hyper-parameter search to optimize the models performance.

Focusing only on the vector operation based fusion strategies was also another limitation of the study.
Bilinear Pooling techniques for fusion has proven to be effective when it comes to identifying relationships.
However, the added computational cost of pooling on top of the transformer module will result in poor performance if training is not done till convergence.
Nonetheless, this is a probable path that can be explored in the future.

Another limitation of the study is that the RAMEN architecture itself may be causing the bottle neck when trying to improve the generalization.
Since both improvements were done to sub-modules of the multi-modal section of the RAMEN model, inherent limitations may exist in the architecture.
Using the knowledge gained from the analysis of the fusion and aggregation module, a new architecture can be developed to well suite domain generalization.

In conclusion, this study paved the path to understanding the characteristics required for domain generalization as well as improving the performance of the RAMEN model.

\section{Acknowledgment}

A special thanks to Robik Shrestha from the Rochester Institute of Technology who was one of the main researchers of the RAMEN study \cite{Shrestha2019} for the guidance and support in providing the links to the datasets and setting up the baseline for this study.

\bibliographystyle{./IEEEtran}
\bibliography{./main}

\onecolumn

\appendices

\section{Total training time}
\label{apdx:trainingtime}
\begin{table} [h]
    \centering
    \begin{tabular}{|l|c|ccc|cccc|} 
    \hline
    \multicolumn{1}{|c|}{\multirow{3}{*}{ \textbf{Dataset } }} & \multicolumn{8}{c|}{\textbf{Training Time per Epoch (minutes) } }                                                         \\ 
    \cline{2-9}
    \multicolumn{1}{|c|}{}                                     & \multicolumn{4}{c|}{\textbf{RAMEN } }                      & \multicolumn{4}{c|}{\textbf{TransformerNet } }               \\ 
    \cline{2-9}
    \multicolumn{1}{|c|}{}                                     & \begin{tabular}[c]{@{}c@{}}\textbf{Baseline}\\\textbf{Concat} \cite{Shrestha2019} \end{tabular} & Additive & Multiplicative & Question Fusion & Concatenation & Additive & Multiplicative & Question Fusion  \\ 
    \cline{1-1}
    \textbf{VQAv1}                                             & 17.00         & 27.51    & 27.67         & 41.52           & 32.75         & 29.04    & 29.10          & 58.12            \\
    \textbf{VQAv2}                                             & 27.15         & 50.80    & 50.94         & 76.59           & 58.45         & 77.88    & 78.30          & 110.40           \\
    \textbf{VQACP2}                                            & 16.29         & 26.27    & 32.74         & 40.08           & 25.00         & 45.39    & 45.67          & 77.67            \\
    \textbf{CVQA}                                              & 11.09         & 14.87    & 15.04         & 26.31           & 21.67         & 42.94    & 28.11          & 39.85            \\
    \textbf{TDIUC}                                             & 48.79         & 85.32    & 86.13         & 129.87          & 91.65         & 89.73    & 90.00          & 183.57           \\
    \textbf{CLEVR}                                             & 28.51         & 49.31    & 51.49         & 47.26           & 28.18         & 56.28    & 56.82          & 63.57            \\
    \textbf{CLEVR-Human}                                       & 0.80          & 1.01     & 1.39          & 1.28            & 0.82          & 1.52     & 92.38          & 2.19             \\
    \textbf{CLEVR-CoGenTA}                                     & 16.40         & 47.47    & 49.28         & 44.95           & 28.11         & 53.52    & 53.74          & 60.49            \\
    \textbf{CLEVR-CoGenTB}                                     & -             & -        & -             & -               & -             & -        & -              & -                \\ 
    \hline
    \multicolumn{1}{|c|}{Average}                              & 20.75         & 37.82    & 39.33         & 50.98           & 35.83         & 49.54    & 59.26          & 74.48            \\ 
    \hline
    \multicolumn{1}{|c|}{Total Average Time}                   & \multicolumn{8}{c|}{\textbf{46.00 } }                                                                                     \\
    \hline
    \end{tabular}
    \end{table}

\section{Training results}
\label{apdx:results}
\begin{adjustbox}{max width=\textwidth}
    \centering
    \begin{tabular}{|l|c|c|c|c|c|c|c|c|c|c|c|c|c|c|c|c|} 
    \hline
    \multicolumn{1}{|c|}{\multirow{3}{*}{ \textbf{ Dataset} }} & \multicolumn{4}{c|}{\textbf{Ramen Concat } }                                                & \multicolumn{4}{c|}{\textbf{TransformerNet Concat } }                                       & \multicolumn{4}{c|}{\textbf{Ramen Additive } }                                              & \multicolumn{4}{c|}{\textbf{TransformerNet Additive } }                                      \\ 
    \cline{2-17}
    \multicolumn{1}{|c|}{}                                     & \multicolumn{2}{c|}{\textbf{Training Score } } & \multicolumn{2}{c|}{\textbf{Test Score } } & \multicolumn{2}{c|}{\textbf{Training Score } } & \multicolumn{2}{c|}{\textbf{Test Score } } & \multicolumn{2}{c|}{\textbf{Training Score } } & \multicolumn{2}{c|}{\textbf{Test Score } } & \multicolumn{2}{c|}{\textbf{Training Score } } & \multicolumn{2}{c|}{\textbf{Test Score } }  \\ 
    \cline{2-17}
    \multicolumn{1}{|c|}{}                                     & Epoch & Score                                  & Epoch & Score                              & Epoch & Score                                  & Epoch & Score                              & Epoch & Score                                  & Epoch & Score                              & Epoch & Score                                  & Epoch & Score                               \\ 
    \hline
    \textbf{VQAv1}                                             & 25    & 88.44                                  & 25    & 63.3                               & 50    & 8388.85                                & 50    & 63.32                              & 25    & 84.65                                  & 25    & 63.21                              & 50    & 58.33                                  & 50    & 55.88                               \\ 
    \hline
    \textbf{VQAv2}                                             & 25    & 84.95                                  & 25    & 62.16                              & 46    & 82.56                                  & 46    & 64.06                              & 25    & 82.43                                  & 25    & 63.64                              & 25    & 62.99                                  & 25    & 59.79                               \\ 
    \hline
    \textbf{VQACP2}                                            & 18    & 80.81                                  & 18    & 37.61                              & 16    & 80.92                                  & 16    & 37.47                              & 14    & 86.53                                  & 14    & 36.73                              & 93    & 53.93                                  & 93    & 27.6                                \\ 
    \hline
    \textbf{CVQA}                                              & 8     & 68.36                                  & 8     & 56.98                              & 16    & 73.61                                  & 16    & 57.74                              & 8     & 68.23                                  & 8     & 55.82                              & 20    & 79.66                                  & 20    & 53.49                               \\ 
    \hline
    \textbf{TDIUC}                                             & -     & -                                      & 9     & 66.48                              & -     & -                                      & 16    & 65.47                              & -     & -                                      & 14    & 64.9                               & -     & -                                      & 7     & 58.03                               \\ 
    \hline
    \textbf{CLEVR}                                             & 20    & 99.75                                  & 20    & 96.52                              & 68    & 96.94                                  & 68    & 95.79                              & 18    & 99.63                                  & 18    & 96.26                              & 16    & 50.5                                   & 16    & 50.52                               \\ 
    \hline
    \textbf{CLEVR-Human}                                       & 31    & 100                                    & 31    & 44.57                              & 77    & 99.2                                   & 77    & 46.49                              & 10    & 87.67                                  & 10    & 40.21                              & 83    & 38.96                                  & 83    & 38.45                               \\ 
    \hline
    \textbf{CLEVR-CoGenTA}                                     & 15    & 99.76                                  & 15    & 96.59                              & 83    & 98.86                                  & 83    & 96.43                              & 24    & 99.82                                  & 24    & 96.84                              & 32    & 64.48                                  & 32    & 64.26                               \\ 
    \hline
    \textbf{CLEVR-CoGenTB}                                     & -     & -                                      & 15    & 88.27                              & -     & -                                      & 83    & 89.68                              & -     & -                                      & 20    & 89.42                              & -     & -                                      & 32    & 55.8                                \\ 
    \hline
    \multicolumn{1}{|c|}{Average}                              & \multicolumn{4}{c|}{68.05}                                                                  & \multicolumn{4}{c|}{68.49}                                                                  & \multicolumn{4}{c|}{67.45}                                                                  & \multicolumn{4}{c|}{51.53}                                                                   \\
    \hline
    \end{tabular}
    \end{adjustbox}

    \begin{adjustbox}{max width=\textwidth}
        \centering
        \begin{tabular}{|l|c|c|c|c|c|c|c|c|c|c|c|c|c|c|c|c|} 
        \hline
        \multicolumn{1}{|c|}{\multirow{3}{*}{\textbf{Dataset }}} & \multicolumn{4}{c|}{\textbf{Ramen Question Fusion }}                                      & \multicolumn{4}{c|}{\textbf{TransformerNet Question Fusion }}                             & \multicolumn{4}{c|}{\textbf{Ramen Multiplicative }}                                       & \multicolumn{4}{c|}{\textbf{TransformerNet Multiplicative }}                               \\ 
        \cline{2-17}
        \multicolumn{1}{|c|}{}                                   & \multicolumn{2}{c|}{\textbf{Training Score }} & \multicolumn{2}{c|}{\textbf{Test Score }} & \multicolumn{2}{c|}{\textbf{Training Score }} & \multicolumn{2}{c|}{\textbf{Test Score }} & \multicolumn{2}{c|}{\textbf{Training Score }} & \multicolumn{2}{c|}{\textbf{Test Score }} & \multicolumn{2}{c|}{\textbf{Training Score }} & \multicolumn{2}{c|}{\textbf{Test Score }}  \\ 
        \cline{2-17}
        \multicolumn{1}{|c|}{}                                   & Epoch & Score                                 & Epoch & Score                             & Epoch & Score                                 & Epoch & Score                             & Epoch & Score                                 & Epoch & Score                             & Epoch & Score                                 & Epoch & Score                              \\ 
        \hline
        \textbf{VQAv1}                                             & 25    & 76.9                                  & 25    & 64.76                             & 25    & 54.61                                 & 25    & 55.08                             & 25    & 70.32                                 & 25    & 65.54                             & 25    & 62.83                                 & 25    & 60.91                              \\ 
        \hline
        \textbf{VQAv2}                                            & 25    & 79.53                                 & 25    & 65.07                             & 25    & 49.31                                 & 25    & 50.14                             & 25    & 68.98                                 & 25    & 65.28                             & 25    & 56.1                                  & 25    & 56.32                              \\ 
        \hline
        \textbf{VQACP2}                                          & 30    & 87.09                                 & 30    & 37.03                             & 26    & 55.75                                 & 26    & 26.89                             & 25    & 71.76                                 & 25    & 36.28                             & 93    & 53.93                                 & 93    & 27.6                               \\ 
        \hline
        \textbf{CVQA}                                            & 8     & 75.38                                 & 8     & 56.81                             & 25    & 56.16                                 & 25    & 48.47                             & 38    & 90.41                                 & 38    & 56.58                             & 17    & 54.2                                  & 17    & 54.2                               \\ 
        \hline
        \textbf{TDIUC}                                           & -     & -                                     & 11    & 64.81                             & -     & -                                     & 11    & 53.9                              & -     & -                                     & 15    & 65.69                             & -     & -                                     & 8     & 56.34                              \\ 
        \hline
        \textbf{CLEVR}                                           & 14    & 97.83                                 & 14    & 96.06                             & 25    & 49.84                                 & 25    & 50.07                             & 21    & 99.86                                 & 21    & 96.72                             & 25    & 57.24                                 & 25    & 57.31                              \\ 
        \hline
        \textbf{CLEVR-Human}                                     & 17    & 99.76                                 & 17    & 48.63                             & 89    & 39.36                                 & 89    & 37.99                             & 25    & 100                                   & 25    & 46.46                             & 25    & 54.64                                 & 25    & 40.07                              \\ 
        \hline
        \textbf{CLEVR-CoGenTA}                                   & 21    & 99.68                                 & 21    & 96.9                              & 23    & 60.35                                 & 23    & 60.24                             & 16    & 99.81                                 & 16    & 96.63                             & 25    & 72.42                                 & 25    & 72.07                              \\ 
        \hline
        \textbf{CLEVR-CoGenTB}                                   & -     & -                                     & 21    & 88.74                             & -     & -                                     & 23    & 55.19                             & -     & -                                     & 16    & 86.22                             & -     & -                                     & 16    & 60.18                              \\ 
        \hline
        \multicolumn{1}{|c|}{Average}                            &       &                                       &       & 68.76                             &       &                                       &       & 48.66                             &       &                                       &       & 68.38                             &       &                                       &       & 53.89                              \\
        \hline
        \end{tabular}
        \end{adjustbox}

\end{document}